\documentclass[journal]{IEEEtran}
\IEEEoverridecommandlockouts

\usepackage{enumitem}
\usepackage[euler]{textgreek}
\usepackage{cite}
\usepackage{amsmath,amssymb,amsfonts}
\usepackage{graphicx}
\usepackage{textcomp}
\usepackage{xcolor}
\usepackage{threeparttable}
\usepackage[hidelinks]{hyperref}
\newcommand{\argmin}{\mathop{\mathrm{argmin}}}
\usepackage{algorithm}
\usepackage{algpseudocode}
\usepackage{booktabs}
\usepackage{siunitx}
\usepackage{multirow}
\usepackage{subfigure}

\usepackage{soul,color}
\soulregister\cite7
\soulregister\ref7
\soulregister\textsubscript7

\graphicspath{{figures/}}

\begin{document}

\title{Few-Shot Load Forecasting Under Data Scarcity in Smart Grids: A Meta-Learning Approach}

\author{
Georgios~Tsoumplekas,
Christos~L.~Athanasiadis, ~\IEEEmembership{Student~Member,~IEEE}, 
Dimitrios~I.~Doukas, ~\IEEEmembership{Member,~IEEE},
Antonios Chrysopoulos,
Pericles~A.~Mitkas, ~\IEEEmembership{Senior~Member,~IEEE}
\thanks{
Georgios~Tsoumplekas is with the Aristotle University of Thessaloniki, Thessaloniki, Greece (e-mail: gktsoump@ece.auth.gr).

Christos~L.~Athanasiadis is with the Department of Electrical and Computer Engineering, Democritus University of Thrace, Xanthi, Greece (e-mail: cathanas@ee.duth.gr) and with NET2GRID BV, Thessaloniki, Greece (e-mail: christos@net2grid.com).

Dimitrios~I.~Doukas and Antonios~Chrysopoulos are with NET2GRID BV, Thessaloniki, Greece (e-mail: dimitrios@net2grid.com, antonios@net2grid.com).

Pericles~A.~Mitkas is with the Department of Electrical and Computer Engineering, Aristotle University of Thessaloniki, Thessaloniki, Greece (e-mail: mitkas@auth.gr).
}

}

\maketitle

\begin{abstract}
    Despite the rapid expansion of smart grids and large volumes of data at the individual consumer level, there are still various cases where adequate data collection to train accurate load forecasting models is challenging or even impossible. This paper proposes adapting an established model-agnostic meta-learning algorithm for short-term load forecasting in the context of few-shot learning. Specifically, the proposed method can rapidly adapt and generalize within any unknown load time series of arbitrary length using only minimal training samples. In this context, the meta-learning model learns an optimal set of initial parameters for a base-level learner recurrent neural network. The proposed model is evaluated using a dataset of historical load consumption data from real-world consumers. Despite the examined load series' short length, it produces accurate forecasts outperforming transfer learning and task-specific machine learning methods by $12.5\%$. To enhance robustness and fairness during model evaluation, a novel metric, mean average log percentage error, is proposed that alleviates the bias introduced by the commonly used MAPE metric. Finally, a series of studies to evaluate the model's robustness under different hyperparameters and time series lengths is also conducted, demonstrating that the proposed approach consistently outperforms all other models.

\end{abstract}


\section{Introduction} \label{introduction}
In recent years, power grids have undergone significant transformations, leading to the emergence of smart grids. Notably, modern power grids have largely shifted away from the traditional paradigm governed by a few big energy producers and multiple smaller consumers. The penetration of renewable energy sources and the appearance of various types of consumers, such as small residential, larger industrial, and even commercial ones (e.g., electric vehicle charging stations), create new challenges\cite{kong2017short}. Consequently, accurate load forecasting, a common desideratum in the energy sector, becomes even more imperative in today's power grids to ensure their security, efficiency, adaptability, and trustworthiness\cite{gasparin2022deep}.

Towards this direction, the widespread adoption of advanced metering infrastructure (AMI) in many countries marks an important milestone. Collecting significant volumes of fine-grained consumption data from individual consumers enables load forecasting at a residential level, unlocking new opportunities for both power providers and load consumers\cite{wang2019probabilistic}. In particular, accurate load forecasting allows for a more precise response to consumer demands, which helps minimize production costs and resource waste while improving the network's reliability during peak load periods. This holds particular significance within the context of modern active distribution networks, characterized by the widespread integration of active assets, i.e., photovoltaic (PV) units and battery energy storage systems, as well as the increasing electrification of traditional fuel-based activities like transportation and heating \cite{athanasiadis2024review}. At the same time, a better understanding of customer profiles allows for better load balancing through the strategic promotion of demand response programs to consumers who are more likely to participate \cite{athanasiadis2024review}.

The abundance of consumers' load consumption data has favored deep learning (DL) techniques for load forecasting in recent years. However, due to the dynamic and ever-changing nature of smart grids, there are various cases where consumption data might be scarce or hard to obtain. Specifically, the increasing penetration of smart grids has led to the addition of newly connected consumers for which historical consumption data is typically sparse. At the same time, meter failures can lead to data corruption, resulting in only a few useful measurements, even for older consumers. Finally, integrating novel assets, such as PV units and heat pumps, can significantly change load consumption patterns, rendering users' historical data immaterial. Under these scenarios, standard DL methods cannot be effectively trained and produce accurate forecasts.

Recently, few-shot learning \cite{tsoumplekas2024toward} (FSL) has emerged as a learning paradigm allowing fast adaptation to novel tasks using only a few training samples. This approach is largely inspired by human learning, where learning a new concept or task is possible given only a few examples or demonstrations\cite{lake2015human}. A common approach to achieve this type of rapid adaptation is using meta-learning\cite{hospedales2021meta} or learning-to-learn. Specifically, meta-learning methods extract information across tasks that share an underlying common structure and leverage it as prior knowledge to adapt to a new task with the same structure without extensive training. Meta-learning techniques in the context of FSL have successfully been applied to various domains, including computer vision\cite{vinyals2016matching, finn2017model} and reinforcement learning\cite{wang2016learning}. Yet their application in load forecasting has only recently been explored \cite{xu2022automated, wang2021clustering, lee2021individualized}.

This paper investigates the potential of meta-learning in the context of FSL for load forecasting. Specifically, the proposed meta-learning model can seamlessly be adapted on the fly to diverse and previously unseen load time series of arbitrary length, which is crucial in cases where a consumer's demand pattern might drastically change or new consumers are added to the smart grid. To achieve this, the proposed method learns a Long Short-Term Memory (LSTM) model's initial parameters to be efficiently optimized using only a few gradient steps within each time series' training set. These initial parameters act as the common knowledge extracted during the meta-learning process. Subsequently, the performance is evaluated using load time series generated by real-world energy consumers with a length of no more than three months. Noticeably, the proposed method significantly outperforms other examined models, producing accurate 15-minute interval energy forecasts for the next seven days. Finally, it manages to maintain consistent performance under varying model hyperparameters and time series lengths.

The contributions of this research can be summarized as follows:

\begin{itemize}

\item A mathematical formulation of meta-learning for time series forecasting is provided, where multiple load consumption time series containing limited measurements from various consumers can be simultaneously used. 

\item The adaptation of Model-Agnostic Meta-Learning++ \cite{antoniou2018train} (MAML++) for the problem of few-shot load forecasting is proposed.

\item A novel evaluation metric, mean absolute log percentage error (MALPE), is proposed that alleviates the bias introduced when using relative errors in model evaluation.

\item The proposed model's performance is evaluated using a set of short-length time series, outperforming transfer learning and task-specific machine learning (ML) approaches by a margin of $12.5\%$.

\item The model's robustness is verified under various hyperparameter settings and time series lengths in ablation studies.

\end{itemize}

\section{Related Work} \label{rel_work}
Since few-shot load forecasting stands at the intersection of load forecasting and FSL, some pertinent work in these areas is discussed before proceeding to existing approaches that combine the two fields.

\subsection{Load Forecasting}
Although there is no universal agreement on the individual time horizons, load forecasting can be split into short-term (a few hours to a few days), mid-term (a month to a year), and long-term (up to several years) forecasting. The proposed method fits within the short-term load forecasting (STLF) framework since it forecasts the load of the next seven days. Due to its aforementioned significance, STLF has been extensively researched in the past few decades, with older approaches including various statistical methods, such as autoregressive moving averages with exogenous variables (ARMAX) models \cite{yang1995identification}. However, during the last two decades, ML and DL techniques have been the primary methods applied for STLF, including support vector machines (SVMs)\cite{mohandes2002support}, convolutional neural networks (CNNs)\cite{kuo2018high}, recurrent neural networks (RNNs)\cite{shi2017deep}, LSTMs \cite{kong2017short}, and attention-based models \cite{sehovac2020deep}. More recently, probabilistic approaches have also been examined\cite{ryu2023quantile}. However, a common characteristic of these approaches is that they use large amounts of data for their training, rendering their application in data-scarce scenarios challenging.

\subsection{Meta-Learning}
The most common approach in FSL is the meta-learning paradigm, which extracts the shared structure among a set of learning tasks and leverages it to generalize to new ones with only a few samples. Meta-learning can generally be classified into (a) metric-based, (b) optimization-based, and (c) model-based meta-learning. Metric-based meta-learning models typically have a backbone that extracts meaningful and transferable representations and an appropriate metric applied to these representations \cite{snell2017prototypical, vinyals2016matching, sung2018learning}. Optimization-based meta-learning relies on learning how to effectively optimize a model to adapt to a new task using only a few training samples, which could include finding a suitable model initialization \cite{finn2017model, antoniou2018train} or learning rates \cite{li2017meta}. Finally, model-based meta-learning employs a meta-learner module that learns how to adapt a base learner to each task \cite{duan2016rl}, and also includes memory-based meta-learning, where an external memory is leveraged to enable faster adaptation of the model to new tasks \cite{santoro2016meta, munkhdalai2017meta}. The proposed method can be classified as an optimization-based meta-learning algorithm, as its objective is to learn a model initialization that can be easily adapted with only a few gradient steps for each task.

\subsection{Few-Shot Load Forecasting}
Traditionally, most of the research related to FSL has been focused on image classification\cite{snell2017prototypical, vinyals2016matching, sung2018learning} and reinforcement learning\cite{duan2016rl, wang2016learning}, while techniques focusing on load forecasting typically assumed the existence of adequate data to approach the problem with task-specific ML methods\cite{wang2019probabilistic}. As a result, the intersection of these two areas has been largely unexplored, with only a limited number of existing works. 

In \cite{xu2022automated}, few-shot load forecasting is cast as a bi-level optimization problem, where in the upper level, the model's hyperparameters are tuned using a tree-based search method, while in the lower level, a meta-learning model similar to that in \cite{finn2017model} is leveraged for forecasting. On the other hand, the authors in \cite{wang2021clustering} perform clustering on the load time series based on a set of extracted features and determine a specific prototype for each cluster. An LSTM model is trained for each prototype and then fine tuned separately on each time series in the cluster. In \cite{lee2021individualized}, a transfer learning model and a meta-learning approach based on MAML \cite{finn2017model} are tested for STLF on individual consumers, achieving promising results. Finally, federated learning has also been leveraged for few-shot hourly load forecasting of different buildings\cite{tang2023privacy} by discriminating buildings into distinct clusters, extracting each cluster's shared structure using an LSTM and fine-tuning a multi-layer perceptron module for each building. Our proposed method bears similarities with the lower-level optimization in \cite{xu2022automated}. However, the robustness to hyperparameter selection achieved by modifying the original MAML obviates the need for an upper-level optimization, thus reducing the computational complexity. At the same time, our proposed method does not involve clustering of tasks, as in \cite{tang2023privacy, wang2021clustering}, further enhancing its computational efficiency.

\section{Methodology} \label{methodology}
This section delineates the proposed method applied for load forecasting under the FSL regime. First, the mathematical formulation of time series forecasting in the context of FSL is rigorously defined. Then, the proposed methodology is described.

\subsection{Problem Formulation} \label{formulation}
In subsequent analysis, the theoretical framework introduced in \cite{brinkmeyer2022few} is adopted and extended to the meta-learning setting for univariate time series forecasting. Specifically, the problem of forecasting a single time series is formalized, following a standard ML approach. This formalization will prove useful later since it will constitute the basis upon which another layer of learning will be added to define the final meta-learning framework of few-shot time series forecasting.

Given a sufficiently large univariate time series, it can be broken down into subsequences that are sequentially fed into an ML model, forming the dataset $D:= \{ (x_j, y_j) \}_{j=1}^N$ of $N$ samples. For that dataset, $x_j \in \mathbb{R}^{T_I}$ constitutes an input subsequence of length $T_I$, while $y_j \in \mathbb{R}^{T_o}$ is the corresponding output subsequence of length $T_o$. Under this setting, the goal of a typical ML approach would be to train a model $f: \mathbb{R}^{T_I} \to \mathbb{R}^{T_o}$, parameterized by $\theta$, to generalize within this dataset. Assuming that there is an underlying joint distribution $p$ that generates this dataset, such that $(x_j, y_j) \sim p(x, y)$ and given an arbitrary loss function $\mathcal{L}: \mathbb{R}^{T_O} \times \mathbb{R}^{T_O} \to \mathbb{R}$, then the optimal set of parameters $\theta^*$ for this model is defined as:

\begin{equation}
    \theta^* = \argmin_{\theta}\mathbb{E}_{(x,y) \sim p(x,y)}[ \mathcal{L}(f(x;\theta), y)]
\end{equation}

Extending this formalization, let us assume that there are $L$ available time series that do not necessarily have the same length. Each one of these time series is characterized by its respective dataset $D_i = \{ (x_{ij}, y_{ij}) \}_{j=1}^{N_i}, i=1,.., L$, where $N_i$ is the number of samples for the $i$-th time series. Similarly to the single time series case, we assume an underlying joint distribution $p_i$ that generates the $i$-th time series' data, such that $(x_{ij}, y_{ij}) \sim p_i(x,y)$. Following the meta-learning terminology, each time series is considered a separate task.  As a consequence, they constitute a set of tasks $\{ \mathcal{T}_i \}_{i=1}^L$ where each one of them is the realization of a task distribution $q$ such that $\mathcal{T}_i \sim q(\mathcal{T})$. Under this setting, the aim of the previously defined model $f$ is to find an optimal set of parameters $\theta^*$ that minimizes the loss across all given tasks. These optimal parameters would, therefore, satisfy:

\begin{equation} \label{eq_a}
    \theta^* = \argmin_{\theta} \mathbb{E}_{\mathcal{T} \sim q(\mathcal{T})} [ \mathbb{E}_{(x,y) \sim p(x,y)} \mathcal{L}(f(x;\theta), y)]
\end{equation}

The existence of the two expectations, one over the task distribution and one over the data distribution within each task renders (\ref{eq_a}) intractable, necessitating its approximation in a tractable form. Since each task corresponds to a specific dataset, we can define a dataset of datasets called \textit{meta-dataset}, defined as $D_{meta}:= \{ D_i \}_{i=1}^L$. The meta-dataset is then split into two disjoint sets, the \textit{meta-train} set $D_{meta-train}$ used for training the model and the \textit{meta-test} set $D_{meta-test}$ used for model evaluation. Next, each task's dataset is split into two disjoint sets: the \textit{support} set $D_i^S$ used for adapting the model parameters within the task and the \textit{query} set $D_i^Q$ used for evaluating the adapted model's performance in that specific task.

To this end, the expected loss over the task distribution can be approximated by the average loss of the meta-train set's tasks. As for the expected loss within each task, the goal is to ensure that the model can generalize strongly within each task given only a few training samples. Even though in the typical empirical risk minimization setting, the loss is minimized on the train set, and the test set is held out for performance evaluation, in the meta-learning regime, both support and query sets are available during training for tasks belonging to the meta-train set. Thus, it makes sense to approximate the expected loss within each task by the average loss over the query sets of those tasks after training the model using the support set samples. Consequently, (\ref{eq_a}) can be approximated by:

\begin{equation} \label{eq_b}
    \theta^* \approx \argmin_{\theta} \left[ \frac{1}{M} \sum_{i=1}^M \left[ \frac{1}{|D_i^Q|} \sum_{(x,y) \in D_i^Q} \mathcal{L}(f(x;\theta|D_i^S), y) \right] \right]
\end{equation} where $M$ is the number of tasks in the meta-train set. The dependence of $\theta$ on the support set in (\ref{eq_b}) will become more apparent in the model description, but briefly, it is the result of fine-tuning the parameters on the support set of the corresponding task.

\subsection{Proposed Method}

The proposed method mainly draws inspiration from MAML++\cite{antoniou2018train}, a meta-learning approach that is a variant of the MAML \cite{finn2017model} algorithm. Before proceeding with the description of the proposed method, it is essential to provide an overview of MAML to understand the underlying logic behind the final model and highlight the reasons for its adoption.

MAML is a gradient-based meta-learning method that aims to generalize within each given task by performing only a few gradient steps on the task's support set. MAML learns an appropriate set of initial parameters to achieve this, granting it this fast adaptation capability. In other words, MAML learns parameters that can be easily fine-tuned within each task. To find this optimal set of initial model parameters, MAML leverages the fact that each task is a realization of a common task distribution and has a shared underlying structure. MAML extracts this transferable knowledge and leverages it to adapt to each task despite the scarcity of training data.

Going back to (\ref{eq_b}), in the context of MAML, $\theta^*$ refers to the optimal set of initial parameters found by minimizing the expression's right-hand side using stochastic gradient descent. The parameter values used to evaluate the model on a task's query set are also formed after fine-tuning the support set. In MAML, this fine-tuning is defined as taking $N_s$ gradient steps on the support set of the task starting from the shared initial parameters. Altogether, MAML can be broken down into two optimization loops or otherwise levels of learning:

\textbf{Inner Loop Optimization (Base Learner):} Given a set of initial parameters $\theta_0$ and a task $\mathcal{T}_i$ with an associated dataset $D_i=\{ D_i^S, D_i^Q \}$, the set of updated parameters is computed by taking $N_s$ gradient steps on the support set $D_i^S$. More specifically, the updated parameters after $k$ gradient steps on $D_i^S$ can be expressed as:

\begin{equation} \label{maml_inner_loop}
    \theta_k^i = \theta_{k-1}^i - \alpha\nabla_{\theta_{k-1}^i}\mathcal{L}_{D_i^S}(f_{\theta_{k-1}^i(\theta_0)})
\end{equation} 

\noindent where $\alpha>0$ is the inner loop learning rate and the dependence of $\theta_{k}^i$ by $\theta_0$ is explicitly denoted and given by unrolling (\ref{maml_inner_loop}).

\textbf{Outer Loop Optimization (Meta-Learner):} Guided by (\ref{eq_b}), the next step is to calculate the mean loss on all of the meta-train set tasks' query sets and optimize the initial parameters by performing stochastic gradient descent using that loss. Specifically, the meta-learner's loss function is defined as:

\begin{equation} \label{meta_loss}
    \mathcal{L}_{meta}(\theta_0) = \sum_{i=1}^M \mathcal{L}_{D_i^Q}(f_{\theta_{N_s}^i(\theta_0)})
\end{equation}

\noindent where $M$ is the total number of training tasks and $f_{\theta_{N_s}^i(\theta_0)}$ is the value of the loss function evaluated on the query set of task $\mathcal{T}_i$ using the fine-tuned parameters that occurred after $N_s$ inner loop optimization steps. The initial model parameters can then be optimized by taking a gradient step using the meta-learner's loss function:

\begin{equation} \label{meta_update}
    \theta_0 = \theta_0 - \beta \nabla_{\theta_0} \mathcal{L}_{meta}(\theta_0)
\end{equation}

\noindent where $\beta>0$ is the outer loop learning rate.

During testing, MAML utilizes the already learned initial parameters and adapts them to the support set of the test task following the same procedure described in the inner loop optimization. After acquiring the adapted parameters, these are utilized for evaluation in the query set.

Despite MAML's success and widespread use in various applications spanning from image classification to reinforcement learning, there are reportedly various challenges associated with using it. These include its training instability, high computational complexity, and the need for careful hyperparameter tuning \cite{antoniou2018train}. To mitigate MAML's aforementioned issues, an updated version of MAML suitable for load forecasting in the context of FSL is proposed. Our method is based on the work of Antoniou et al.\cite{antoniou2018train}, which extends the MAML architecture and has the following main features:

\textbf{Multi-Step Loss Function.} To achieve stability during training, optimization based on gradients of all inner loop steps is required. In MAML, the gradient step on the outer loop is performed using the base learner's loss value on a task's query set after the inner loop gradient steps have been performed, as seen in (\ref{meta_loss}). However, failing to consider the gradients during the intermediate steps explicitly is one of the main causes of instability since it hinders smooth gradient propagation in the network. By leveraging an outer loop loss function that is a weighted average of the query set losses of every inner loop gradient step, we ensure that gradients from all inner loop steps are included both explicitly (through the current step's query set loss) and implicitly (by unrolling the inner loop during subsequent steps) in the outer loop optimization procedure. More specifically, this multi-step loss is defined as:

\begin{equation} \label{multi_step_loss}
    \mathcal{L}_{meta}^{MS}(\theta_0) = \sum_{i=1}^M \sum_{k=1}^{N_s} v_k \mathcal{L}_{D_i^Q}(f_{\theta_k^i(\theta_0)})
\end{equation}

\noindent where $\mathcal{L}_{D_i^Q}(f_{\theta_k^i(\theta_0)})$ is the base learner's query set loss on a training task $D_i \in D_{meta-train}$ after $k$ inner loop gradient steps and $v_k$ is the loss weight during that step. 

It must be highlighted that to ensure stability during the early stages of training, all loss weight values are set to be close to each other. However, as training progresses, it is crucial to prioritize the last gradient step's loss to enhance the effect of minimizing the loss after the last step of the inner loop. The rationale behind this decision stems from the fact that query set evaluation is performed using the model parameters after that step during testing. A weight matrix can be defined, $\mathbf{V} \in \mathbb{R}^{N_e \times N_s}$, where $N_e$ is the number of training epochs, and $v_{e,k} \in \mathbf{V}$ corresponds to the loss weight after the $k^{th}$ inner loop gradient step during the $e^{th}$ training epoch. A training epoch is equivalent to one outer-loop optimization step in the proposed method. During the first inner loop gradient step, all loss weights are equal:

\begin{equation} \label{weight_matrix_1}
    v_{1,k} = \frac{1}{N_s}
\end{equation}

\noindent In the subsequent steps, the loss weights of intermediate steps are reduced following a method similar to simulated annealing. In contrast, the loss weight of the last step is analogously increased. The rest of matrix $\mathbf{V}$ can be defined as:

\begin{equation} \label{weight_matrix_2}
    v_{e,k} = 
        \begin{cases}
            \max \{v_{e-1,k} - \frac{e}{N_s^2}, \frac{\gamma}{N_s}\}, \ \ \ \  k=2,3,..,N_s-1 \\
            \min \{v_{e-1,k}+\frac{e(N_s-1)}{N_s^2}, 1-\frac{\gamma(N_s-1)}{N_s}\}, \ \ k=N_s
        \end{cases}
\end{equation}

\noindent where $\frac{1}{N_s^2}$ is the decay rate, $0 < \gamma < 1$, and $\frac{\gamma}{N_s}$ is the maximum value of the intermediate steps' loss weights. From the above definition, $\sum_{k=1}^{N_s} v_{e,k} = 1$, and so the role of these values as weights for the calculation of a weighted average loss is apparent. Given that the inner loop steps and the total number of training epochs are known beforehand, $\mathbf{V}$ can be calculated for the whole training procedure, hence not imposing any additional computational burden during each epoch. In (\ref{multi_step_loss}), the $e$ indices have been omitted for simplicity.

\textbf{Higher-order derivatives computation annealing.} One of the simplest ways to reduce MAML's computational complexity is by omitting the calculation of second-order derivatives. This simplification was initially suggested by Finn et al.\cite{finn2017model}, who empirically showed that this approximation leads to speedups up to $33\%$ without any significant performance drop. However, these higher-order gradients typically carry rich information that could enhance the meta-learning model's generalization capabilities. A training procedure that leverages both first and second-order derivatives is introduced to ensure an effective compromise between performance and computational efficiency. During the first training epochs, a first-order approximation of second-order derivatives is utilized to speed up training. In contrast, the second-order derivatives are calculated and used to enhance model performance during the remaining epochs.

\textbf{Learnable Per-Layer and Per-Step Inner Loop Learning Rates.} To alleviate the need for a thorough search for optimal hyperparameters that would ensure the model's stability, Antoniou et al.\cite{antoniou2018train} propose a modification that constitutes a compromise between MAML's need for extensive hyperparameter tuning and MetaSGD's\cite{li2017meta} additional computational burden due to meta-learning a different inner loop learning rate for each model parameter. Specifically, they propose that instead of learning a separate learning rate for every parameter, parameters in the same layer can share a single meta-learnable learning rate. The fact that the model can learn its learning rates induces robustness since it can readjust them properly without additional human intervention. At the same time, compared to MetaSGD, the computational overhead is significantly decreased since the number of layers is much smaller than the total number of parameters, even in deep architectures. 

Moreover, to reduce the risk of overfitting, the meta-learner outputs different learning rates for each step of the base learner's optimization in the inner loop. Combining all of the above, let us consider a base learner with $L$ layers and $P$ parameters per layer. For that base learner, MAML would consist of $L \times P$ parameters, MetaSGD of $2 \times L \times P$ parameters, and MAML++ of $L \times (P + N_s-1)$ parameters. Typically, $N_s \ll P$ and, as a result, MAML++ enjoys much of MetaSGD's flexibility and robustness while imposing only a minor computational overhead to the original MAML architecture.

\textbf{Cosine Annealing of Outer Loop Learning Rate.} To facilitate the model's faster convergence and to enhance generalization across tasks, cosine annealing\cite{loshchilov2016sgdr} is employed to determine the outer loop's learning rate. Using cosine annealing also increases the model's robustness to different hyperparameter values since it removes the need to carefully select a static outer learning rate that will not lead to instability during training. The meta-learner's learning rate is given by:

\begin{equation} \label{cosine_annealing}
    \beta_e = \beta_{min} + \frac{1}{2}(\beta_{max} - \beta_{min})(1 + cos\frac{e}{N_{e_{max}}}\pi)
\end{equation}

\noindent where $\beta_e \in [\beta_{min},\beta_{max}]$ is the outer loop learning rate during the $e^{th}$ epoch and $N_{e_{max}}$ is the total number of epochs before resetting the learning rate's value. Consequently, by combining (\ref{meta_update})-(\ref{multi_step_loss}) with (\ref{cosine_annealing}), the meta-learner's update step is given by:

\begin{equation} \label{MAML++outer}
    \theta_0 = \theta_0 - \beta_e \nabla_{\theta_0} \mathcal{L}_{meta}^{MS}(\theta_0)
\end{equation}

\begin{algorithm}[h!]
\caption{Meta-learning model training}\label{alg:1}
\begin{algorithmic}[1]
    \Require Meta-train set $D_{meta-train}$
    \Require Number of training epochs $N_e$
    \Require Number of inner loop steps $N_s$
    \State Calculate weight matrix $\mathbf{V}$ using (\ref{weight_matrix_1})-(\ref{weight_matrix_2})
    \State Randomly initialize $\theta_0$
    \For{$epoch \gets 1,2,..,N_e$}
        \For{$D_i \in D_{meta-train}$}
            \State Sample support set $D^S_i$
            \For{$step_k \gets 1,2,...,N_s$}
                \State Compute $\mathcal{L}_{D_i^S}(f_{\theta^i_{step_{k-1}}(\theta_0)})$
                \State Update $\theta^i_{step_k}$ using (\ref{maml_inner_loop})
            \EndFor
            \State Sample query set $D^Q_i$
            \State Compute $\mathcal{L}_{D_i^Q}(f_{\theta^i_{step_{N_s}}(\theta_0)})$
        \EndFor
        \State Compute $\mathcal{L}_{meta}^{MS}(\theta_0)$ using (\ref{multi_step_loss})
        \State Compute $\beta_{epoch}$ using (\ref{cosine_annealing})
        \State Update $\theta_0$ using (\ref{MAML++outer})
    \EndFor    
\end{algorithmic}
\end{algorithm}

The whole training procedure of the proposed method is outlined in Algorithm \ref{alg:1}. Regarding the corresponding evaluation procedure, it is much faster and more lightweight since the initial model parameters have already been learned during training, obfuscating the use of the outer loop during testing. More specifically, the model is initialized for each task using the meta-learned initial parameters, which are then adapted to the task's support set by performing $N_s$ gradient steps. The corresponding inner loop learning rate per layer that has been meta-learned during training is utilized for each step. Finally, based on these adapted parameters, the model's performance is evaluated on the task's query set.

\section{Experiments} \label{experiments}
This section evaluates the proposed method's ability to adapt quickly and produce accurate forecasts for new load time series under various FSL settings. More specifically, the experimental setup includes the dataset, the experiment and model settings, the baselines, and the evaluation metrics. Next, the main findings that demonstrate the superiority of our method are presented, and the method's performance is also evaluated for time series of different lengths. Finally, we conduct a series of ablation studies, to assess the robustness of the proposed method under various architectural design choices.

\subsection{Experimental Setup}
\textbf{Dataset.} A set of univariate load time series from European load consumers was used for the experiments. Each time series is the aggregate load of 50 individual consumers, consists of 15-minute interval energy measurements, and was collected from October 1, 2020, to April 1, 2022. Each time series constitutes a separate task; thus, for the rest of this paper, time series and task are used interchangeably. The meta-train set contains 40 tasks of length between two and four months, starting from the first day of any of the months in the examined period. The meta-test set contains 15 time series starting from each month in the total examined period, with five each having a length of one, two, and three months, resulting in a total of 240 test tasks.

Longer time series were used during meta-training than meta-testing, emulating the real-world scenario where an energy provider trains a meta-learning model on energy consumption data from older clients and tests it on new clients with limited available data. Due to their short length, individual time series cannot encapsulate seasonality. Nonetheless, the basic hypothesis is that the proposed model can extract the necessary knowledge to learn the underlying seasonal component by training on a set of time series that will cover all months of the calendar year.

\begin{figure}[h]
\includegraphics[scale=0.25]{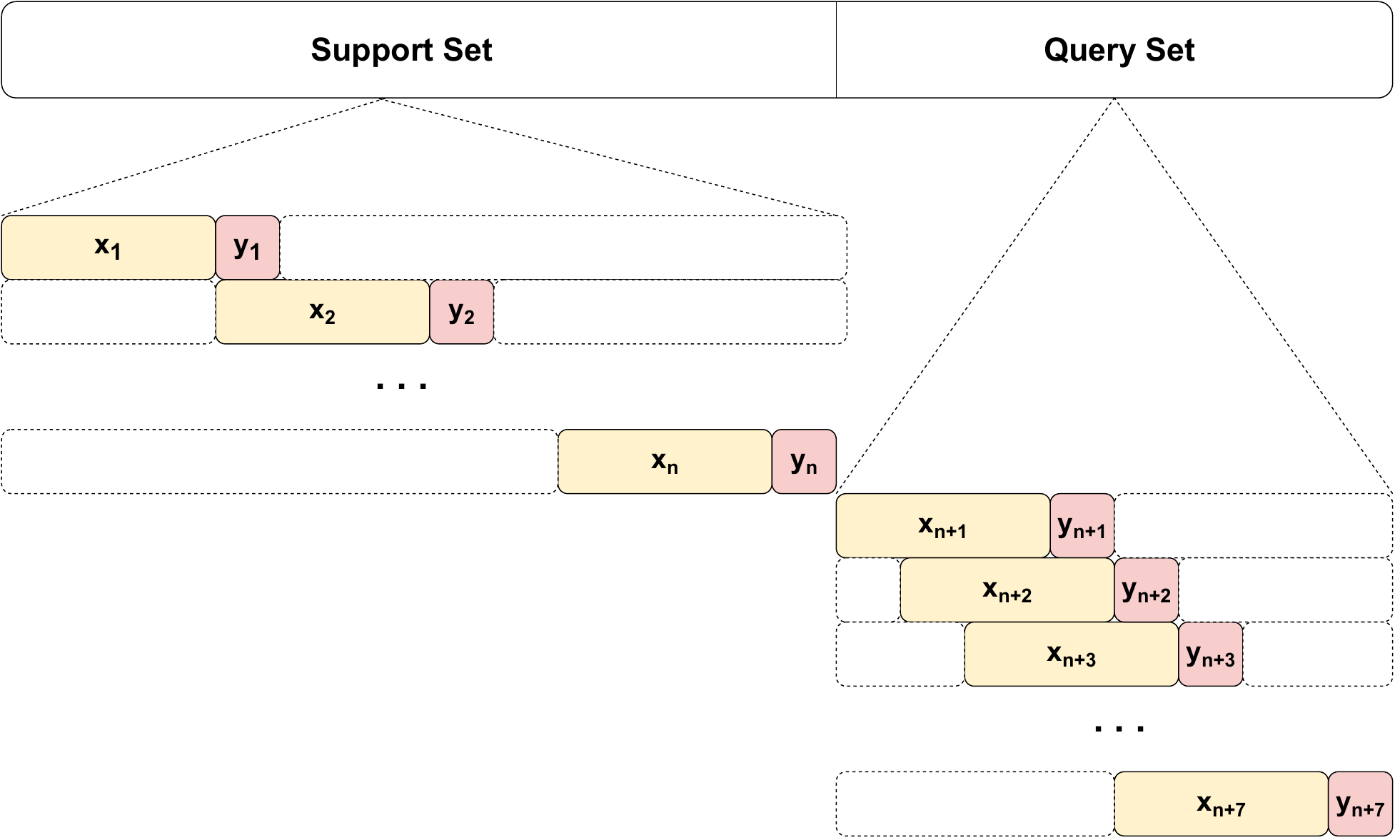}
\caption{Each task's support set has a variable size and non-overlapping samples, while the query set has a fixed size and overlapping samples with a 1-day shift.}
\label{support_query_split}
\end{figure}

The problem is formulated as a sequence-to-sequence forecasting problem. Each sample consists of measurements taken over a week; the corresponding outputs are the measurements for the following day. Support set samples are obtained by splitting the part of the time series that belongs to the support set into samples with non-overlapping input sequences. However, for query set samples, we use overlapping input sequences, shifted by one day, to get more samples from a smaller part of the time series. Fig. \ref{support_query_split} illustrates these differences in the sample splitting process, where $x_i$ is a sample's input sequence and $y_i$ is its corresponding output sequence. To allow for a fair evaluation across different tasks, all query sets have a fixed length of seven samples, each with an output sequence corresponding to a different day of the week. The support sets have varying lengths, allowing for training with a different number of samples (shots). The difference between the number of samples in the support and query sets is also illustrated in Fig. \ref{support_query_split}, where a task has $n$ support set samples and seven query samples.

\begin{figure*}
  \centering
  \subfigure[MSE Boxplots]{
    \includegraphics[scale=0.74]{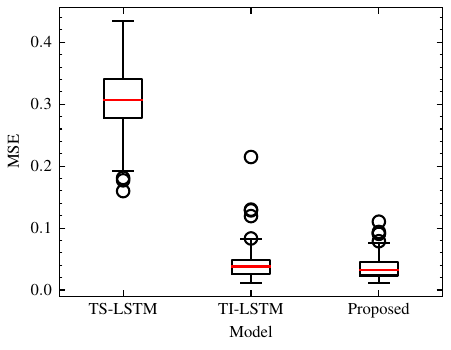}
    \label{main_results_mse}}
  \subfigure[MAPE Boxplots]{ 
   \includegraphics[scale=0.74]{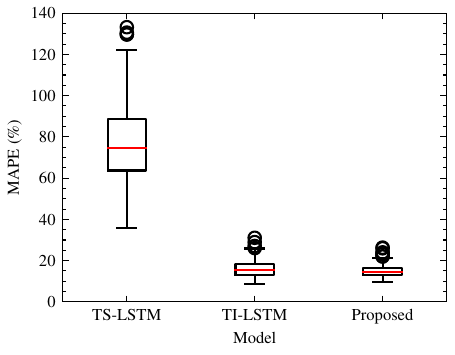}
    \label{main_results_mape}}
  \subfigure[MALPE Boxplots]{
  \includegraphics[scale=0.74]{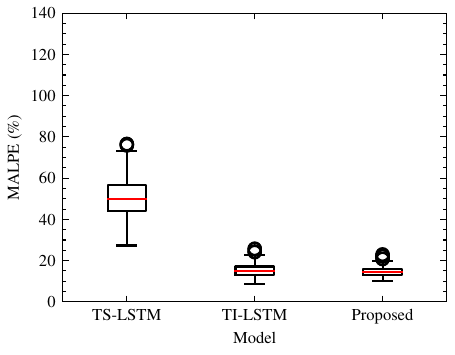}
    \label{main_results_malpe}}
  \caption{Boxplots of the reported model performance metrics for the main experiment setting. The proposed model produces more accurate and robust results than the baseline methods.}
  \label{main_results_boxplots}
\end{figure*}

\begin{figure*}[t]
\includegraphics[width=\textwidth]{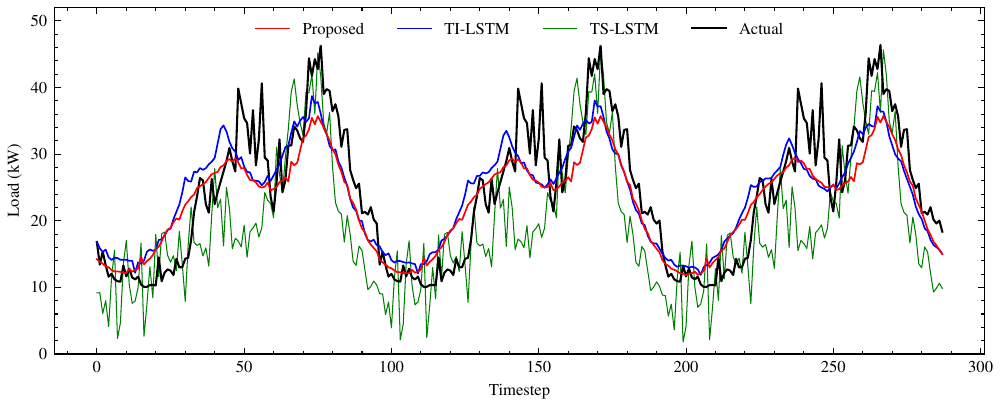}
\caption{Model forecasts on the first three days of a meta-test set task's query set.}
\label{main_experiment_forecasts}
\end{figure*}

\textbf{Baseline Models.} Following \cite{iwata2020few}, two different baselines corresponding to two different training frameworks were introduced: (a) a task-specific and (b) a task-independent learning framework. Task-specific learning refers to the standard ML approach, where a separate model is trained for each task. On the other hand, task-independent learning involves training a single model using all data available in the meta-train set. Consequently, the model is fine-tuned on the support set and evaluated on the query set for each testing task. This is similar to transfer learning, where the combination of the meta-training tasks corresponds to the source task, and each meta-testing task is the target task. For brevity, the task-specific model will be referred to as TS-LSTM, and the task-independent model as TI-LSTM.

\textbf{Experimental Details.} To ensure a fair comparison, all three models use the same LSTM backbone consisting of an LSTM layer followed by a linear layer with 32 hidden units. The number of inner loop training steps for the proposed method, the number of fine-tuning epochs for TI-LSTM, and the number of training epochs for TS-LSTM are all set to 1. The proposed method and TI-LSTM are trained for 150 epochs on the meta-train set. Finally, the proposed method uses first-order gradient approximation for the first 50 epochs.

\textbf{Evaluation Metrics.} For the evaluation of the proposed method and baselines, both Mean Squared Error (MSE) and Mean Absolute Percentage Error (MAPE) are reported. MSE is also used as the base loss function in the optimization process. Moreover,  a novel evaluation metric is introduced, the Mean Absolute Log Percentage Error (MALPE), which is defined as follows:

\begin{equation}
    MALPE = \frac{100\%}{N} \sum_{i=1}^N \left|\log\left(\frac{\hat{y}_i}{y_i}\right)\right|
\end{equation}

$\hat{y}_i$ is the forecast value, $y_i$ is the corresponding actual value, and $N$ is the number of query set samples in each task. The proposed metric replaces the relative error in MAPE with the logarithm of the predicted to true values ratio. MALPE is symmetric, and its formulation is derived following \cite{tofallis2015better} where the use of the logarithm is theoretically justified to mitigate the bias introduced by the relative error. Specifically, evaluating models on MAPE leads to biased results since MAPE favors underforecasting models and strictly penalizes overforecasting ones. Finally, it is worth noting that since we refer to a set of testing tasks, all metrics are calculated for each task separately, and the mean and standard deviation values are reported.

\subsection{Main Results}

The results obtained for the main experiment setting are summarized in Fig. \ref{main_results_boxplots}. Specifically, the proposed method and the baselines are evaluated in each of the 240 meta-test set tasks, and the corresponding MSE, MAPE, and MALPE boxplots are created for each model. Our proposed model outperforms both baselines in all three metrics, improving MSE by $12.5\%$ compared to TI-LSTM ($0.035$ vs $0.040$), the second-best performing model. Additionally, the proposed model demonstrates increased performance robustness for different tasks, as the results are more concentrated towards the median reported value than the other two models. 

Interestingly, the performance gap between the proposed method and TI-LSTM is smaller than that between TI-LSTM and TS-LSTM, outlining the importance of incorporating prior knowledge to quickly adapt to novel tasks with few samples. At the same time, explicitly incorporating prior knowledge in the model as an inductive bias (in the proposed method as optimal initial parameters) is more effective than implicit incorporation through model pretraining. The fact that our meta-learning method manages to outperform TI-LSTM is particularly encouraging, especially in the light of various recent studies highlighting the strong performance of fine-tuning-based models for FSL problems\cite{chen2019closer, tian2020rethinking}.

\begin{table*}[t]
\centering
\caption{Monthly model performance for the examined period (mean reported values averaged over all 3-month meta-test set tasks)}
\label{per_month_results_table}
\noindent\resizebox{\textwidth}{!}{\begin{tabular}
{m{1.5cm}m{1.5cm}SSSSSSSSSSSSSS}
\toprule
{\textbf{Model}} & {\textbf{Metric}} & {\textbf{Feb-21}} & {\textbf{Mar-21}} & {\textbf{Apr-21}} & {\textbf{May-21}} & {\textbf{Jun-21}} & {\textbf{Jul-21}} & {\textbf{Aug-21}} & {\textbf{Sep-21}} & {\textbf{Oct-21}} & {\textbf{Nov-21}} & {\textbf{Dec-21}} & {\textbf{Jan-22}} & {\textbf{Feb-22}} & {\textbf{Mar-22}} \\
\midrule
TS-LSTM & {\multirow{3}{*}{MSE}} & {0.276} & {0.342} & {0.372}  & {0.343} & {0.368} & {0.340} & {0.277} & {0.292} & {0.272} & {0.278} & {0.216} & {0.248} & {0.304} & {0.324} \\
TI-LSTM & & {0.047} & {0.057} & {0.017} & {\textbf{0.018}} & {0.033} & {\textbf{0.018}} & {0.038} & {\textbf{0.028}} & {\textbf{0.029}} & {0.062} & {0.042} & {0.047} & {0.031} & {0.031} \\
Proposed & & {\textbf{0.029}} & {\textbf{0.039}} & {\textbf{0.016}}  & {0.021} & {\textbf{0.024}} & {0.020} & {\textbf{0.033}} & {0.030} & {0.034} & {\textbf{0.038}} & {\textbf{0.034}} & {\textbf{0.037}} & {\textbf{0.024}} & {\textbf{0.024}} \\
\midrule
TS-LSTM & {\multirow{3}{*}{MAPE}} & {78.80\%} & {97.65\%} & {103.72\%}  & {99.14\%} & {98.39\%} & {86.10\%} & {69.81\%} & {69.10\%} & {64.97\%} & {66.19\%} & {54.36\%} & {63.67\%} & {86.58\%} & {86.66\%} \\
TI-LSTM & & {19.86\%} & {24.61\%} & {13.16\%} & {\textbf{13.54\%}} & {18.27\%} & {\textbf{13.81\%}} & {15.94\%} & {\textbf{12.95\%}} & {\textbf{12.17\%}} & {15.90\%} & {15.58\%} & {16.78\%} & {17.16\%} & {16.49\%} \\
Proposed & & {\textbf{15.50\%}} & {\textbf{19.11\%}} & {\textbf{12.72\%}} & {15.01\%} & {\textbf{15.96\%}} & {14.72\%} & {\textbf{15.87\%}} & {13.79\%} & {12.93\%} & {\textbf{12.76\%}} & {\textbf{14.22\%}} & {\textbf{14.27\%}} & {\textbf{14.61\%}} & {\textbf{14.17\%}} \\
\midrule
TS-LSTM & {\multirow{3}{*}{MALPE}} & {49.98\%} & {58.50\%} & {64.77\%}  & {61.64\%} & {62.19\%} & {57.86\%} & {46.94\%} & {47.42\%} & {45.75\%} & {44.77\%} & {38.88\%} & {42.79\%} & {55.52\%} & {56.54\%} \\
TI-LSTM & & {17.68\%} & {21.09\%} & {12.60\%} & {\textbf{13.38\%}} & {16.60\%} & {\textbf{13.14\%}} & {15.52\%} & {\textbf{12.70\%}} & {\textbf{12.32\%}} & {16.30\%} & {15.13\%} & {15.72\%} & {15.93\%} & {15.04\%} \\
Proposed & & {\textbf{14.27\%}} & {\textbf{17.00\%}} & {\textbf{12.50\%}} & {14.57\%} & {\textbf{15.14\%}} & {14.14\%} & {\textbf{15.39\%}} & {13.89\%} & {13.26\%} & {\textbf{13.07\%}} & {\textbf{14.26\%}} & {\textbf{13.87\%}} & {\textbf{14.15\%}} & {\textbf{13.21\%}} \\
\bottomrule
\end{tabular}}
\end{table*}

\begin{table*}[t]
\centering
\caption{Mean model performance for varying time series support set lengths}
\label{support_set_size_table}
\noindent\resizebox{\textwidth}{!}{\begin{tabular}
{m{1.5cm}m{1.5cm}SSSSSSSSSSSS}
\toprule
\multirow{2}{*}{\textbf{Model}} & \multirow{2}{*}{\textbf{Metric}} & \multicolumn{12}{c}{\textbf{Support set size (months)}} \\
    \cmidrule(r){3-14}
    & & {\textbf{1}} & {\textbf{2}} & {\textbf{3}} & {\textbf{4}} & {\textbf{5}} & {\textbf{6}} & {\textbf{7}} & {\textbf{8}} & {\textbf{9}} & {\textbf{10}} & {\textbf{11}} & {\textbf{12}} \\
\midrule
TS-LSTM & {\multirow{3}{*}{MSE}} & {$0.308$} & {$0.312$} & {$0.306$}& {$0.294$} & {$0.280$} & {$0.261$} & {$0.224$} & {$0.174$} & {$0.122$} & {$0.090$} & {$0.083$} & {$0.060$} \\
TI-LSTM & & {$0.048$}& {$0.037$}& {$0.035$}& {$0.031$} & {$0.030$} & {$0.029$} & {$\mathbf{0.029}$} & {$\mathbf{0.028}$} & {$\mathbf{0.027}$} & {$\mathbf{0.025}$} & {$\mathbf{0.024}$} & {$\mathbf{0.025}$} \\
Proposed & & {$\mathbf{0.045}$}& {$\mathbf{0.032}$}& {$\mathbf{0.028}$}& {$\mathbf{0.026}$} & {$\mathbf{0.026}$} & {$\mathbf{0.027}$} & {$0.033$} & {$0.034$} & {$0.036$} & {$0.037$} & {$0.038$} & {$0.043$} \\
\midrule
TS-LSTM & {\multirow{3}{*}{MAPE}} & {$72.07$} & {$78.26$} & {$80.82$} & {$83.39$} & {$83.21$} & {$81.41$} & {$78.18$} & {$64.78$} & {$49.91$} & {$37.41$} & {$31.60$} & {$25.98$} \\
TI-LSTM & & {$16.22$} & {$16.05$} & {$15.94$} & {$16.01$} & {$15.48$} & {$\mathbf{15.75}$} & {$\mathbf{16.51}$} & {$\mathbf{16.32}$} & {$\mathbf{15.74}$} & {$\mathbf{15.79}$} & {$\mathbf{15.19}$} & {$\mathbf{15.09}$} \\
Proposed & & {$\mathbf{15.75}$} & {$\mathbf{14.61}$} & {$\mathbf{14.58}$} & {$\mathbf{15.19}$} & {$\mathbf{15.34}$} & {$16.31$} & {$19.45$} & {$19.14$} & {$20.34$} & {$21.05$} & {$20.61$} & {$21.96$} \\
\midrule
TS-LSTM & {\multirow{3}{*}{MALPE}} & {$47.98$} & {$51.35$}  & {$52.73$} & {$54.11$}  & {$54.23$}  & {$53.65$}  & {$51.98$}  & {$45.30$} & {$38.37$} & {$33.47$} & {$31.29$} & {$27.90$} \\
TI-LSTM & & {$15.69$} & {$15.25$} & {$15.04$} & {$15.19$}& {$14.94$} & {$\mathbf{14.82}$} & {$\mathbf{15.54}$} & {$\mathbf{15.52}$} & {$\mathbf{15.10}$} & {$\mathbf{14.85}$} & {$\mathbf{14.48}$} & {$\mathbf{14.53}$} \\
Proposed & & {$\mathbf{15.67}$} & {$\mathbf{14.24}$} & {$\mathbf{14.16}$} & {$\mathbf{14.73}$} & {$\mathbf{14.92}$} & {$15.80$} & {$18.89$} & {$18.48$} & {$19.64$} & {$21.14$} & {$20.49$} & {$21.69$} \\
\bottomrule
\end{tabular}}
\end{table*}

The forecasts of the baselines and the proposed method on the first three days of a meta-test task's query set with a total length of 1 month are illustrated in Fig. \ref{main_experiment_forecasts}. The three days to be forecast in the query set can be seen by the periodic patterns of the actual time series. Following the results presented in Fig. \ref{main_results_boxplots}, the proposed meta-learning method and TI-LSTM produce more accurate forecasts than TS-LSTM, whose forecasts are discerned by large fluctuations and deviations from the actual values. On the other hand, while the disparity between the proposed method and TI-LSTM is subtle, it is evident that the former model produces smoother results that better match the actual time series.

Finally, Table \ref{per_month_results_table} contains the mean monthly metric values for each model averaged over all meta-test set tasks for 3 months. Each task in this setting consists of a 3-month support set, followed by a 1-month query set corresponding to a different month of the year. The proposed model outperforms TI-LSTM for 10 out of 14 examined months while being marginally inferior to TI-LSTM for the remaining months due to their high variability, slightly favoring TI-LSTM's predictions. The proposed method's consistent performance across all months substantiates our hypothesis that it can learn the underlying seasonality even when trained on sub-seasonal time series.

\subsection{Effect of support set size}

This study investigates how varying the size of each task's support set affects model performance. It is expected to be more challenging for models to generalize based on tasks with fewer samples since there is less data to which they can adapt. At the same time, moving further away from the few-shot regime toward time series with more data, it is anticipated that the gap between meta-learning and task-specific approaches will become smaller.

Table \ref{support_set_size_table} provides the model performance metric values for a 1-month query set when the support set ranges from one to 12 months. As anticipated, the proposed model outperforms both baseline models when the support set size of the new input tasks does not exceed six months (five for MAPE and MALPE metrics), verifying meta-learning's effectiveness under data scarcity. By analyzing the MSE values, it is clear that our proposed model attains an MSE of 0.028 with just three months of training data. In contrast, the TI-LSTM model reaches a comparable error level only after training with eight months of historical data. This implies that our proposed method achieves similar performance with five months less data pinpointing its practical efficacy in data scarce settings. However, when the support set length exceeds a limit, our model starts underfitting the data provided, emphasizing the need for additional inner loop steps.

On the other hand, both TI-LSTM's and TS-LSTM's performance improves when increasing the support set size, but only TI-LSTM manages to outperform the proposed method when moving further away from the few-shot data regime. Nonetheless, it is evident that leveraging prior knowledge to improve model performance and facilitate adaptation to new tasks is crucial in both data regimes and becomes increasingly important when training samples are extremely scarce (e.g., 32.1\% improvement in MALPE between our method and TS-LSTM for support set size of one month vs 6.21\% improvement for support set size of 12 months).

\subsection{Ablation Study}

This subsection reports the results of a series of experiments that aim to shed further light on the proposed method's performance. By altering various experimental and model configurations, the proposed method's robustness under these variations and its efficacy under scenarios of extreme data scarcity are demonstrated.

\begin{table}[h]
\centering
\caption{Proposed model performance for different number of training epochs using second-order derivatives}
\label{second_order_table}
\noindent\resizebox{\columnwidth}{!}{\begin{tabular}{m{2.5cm}SSS}
\toprule
{\textbf{Model}} & {\textbf{MSE}} & {\textbf{MAPE}} & {\textbf{MALPE}} \\
\midrule
Proposed (s.o = 0) & {$0.047_{\pm0.029}$} & {$16.18\%_{\pm3.38\%}$} & {$17.18\%_{\pm4.23\%}$} \\
Proposed (s.o = 50) & {$0.036_{\pm0.017}$} & {$\mathbf{14.78\%_{\pm2.79\%}}$} & {$\mathbf{14.67\%_{\pm2.48\%}}$} \\
Proposed (s.o = 100) & {$\mathbf{0.035_{\pm0.017}}$} & {$15.00\%_{\pm2.92\%}$} & {$14.72\%_{\pm2.52\%}$} \\
Proposed (s.o = 150) & {$0.037_{\pm0.017}$} & {$15.80\%_{\pm3.46\%}$} & {$15.13\%_{\pm2.77\%}$} \\
\bottomrule
\end{tabular}}
\end{table}

\begin{table*}
\centering
\caption{Model performance for different number of linear layers}
\label{num_layers_table}
\noindent\resizebox{\textwidth}{!}{\begin{tabular}{m{1.5cm}SSSSSSSSS}
    \toprule
    \multirow{2}{*}{\textbf{Model}} & \multicolumn{3}{c}{\textbf{1 layer}} & \multicolumn{3}{c}{\textbf{2 layers}} & \multicolumn{3}{c}{\textbf{3 layers}} \\
    \cmidrule(r){2-4}\cmidrule(l){5-7}\cmidrule(l){8-10}
    & {\textbf{MSE}} & {\textbf{MAPE}} & {\textbf{MALPE}} & {\textbf{MSE}} & {\textbf{MAPE}} & {\textbf{MALPE}} & {\textbf{MSE}} & {\textbf{MAPE}} & {\textbf{MALPE}} \\
    \midrule
    {TS-LSTM} & {$0.309_{\pm0.052}$} & {$76.87\%_{\pm18.73\%}$} & {$50.59\%_{\pm9.55\%}$} & {$0.324_{\pm0.053}$} & {$78.63\%_{\pm18.91\%}$} & {$51.20\%_{\pm9.49\%}$} & {$0.330_{\pm0.056}$} & {$79.37\%_{\pm19.23\%}$} & {$51.60\%_{\pm9.66\%}$} \\
    TI-LSTM & {$0.040_{\pm0.022}$} & {$16.07\%_{\pm4.02\%}$} & {$15.34\%_{\pm3.20\%}$} & {$0.042_{\pm0.023}$} & {$16.61\%_{\pm4.11\%}$} & {$15.86\%_{\pm3.35\%}$} & {$\mathbf{0.041_{\pm0.018}}$} & {$\mathbf{17.63\%_{\pm4.70\%}}$} & {$\mathbf{16.31\%_{\pm3.59\%}}$} \\
    \midrule
    Proposed & {$\mathbf{0.035_{\pm0.017}}$} & {$\mathbf{15.00\%_{\pm2.92\%}}$} & {$\mathbf{14.72\%_{\pm2.52\%}}$} & {$\mathbf{0.034_{\pm0.016}}$} & {$\mathbf{15.88\%_{\pm3.55\%}}$} & {$\mathbf{15.12\%_{\pm2.93\%}}$} & {$0.047_{0.016}$} & {$21.07\%_{\pm7.30\%}$} & {$18.97\%_{\pm5.27\%}$} \\
    \bottomrule
\end{tabular}}
\end{table*}

\textbf{Second Order Derivatives.} In the first of these experiments, the varying number of training epochs incorporating second-order gradients in the optimization process impact on the proposed model's performance is investigated. Table \ref{second_order_table} presents the mean and standard deviation metric values on the meta-test set when training the model for a different number of epochs with second-order derivatives activated (denoted as s.o.). Using only first-order approximation decreases performance, suggesting that second-order gradients carry essential information for generalizing to unknown tasks. However, using second-order derivatives throughout all training epochs also harms model performance, possibly due to overfitting since the number of training tasks is limited. As a result, combining both approaches is optimal. The proposed model uses second-order gradients for the main experiment and subsequent ablation studies for the last 100 training epochs. This leads to the best MSE and second-best MAPE and MALPE results across all examined variations.

\textbf{Base Learner Depth.} Next, the impact of the base model depth on performance is examined. Based on the work in \cite{arnold2021maml}, gradient-based meta-learning models, like the one proposed, can benefit from deeper architectures since the early layers of these networks extract the task representations (task representation learning). In contrast, the final layers share the extracted knowledge across tasks through meta-learning. As a result, adding layers before the output layer can theoretically facilitate meta-learning. To ensure that possible performance changes can be attributed solely to increased model depth, all added layers are linear (without any nonlinear activation functions) to not increase model capacity. Specifically, the number of linear layers gradually increases to two and three while the rest of the experiment and model settings remain the same.

In Table \ref{num_layers_table}, it can be observed that the proposed method's performance is minimally affected when the number of linear layers is increased to two, maintaining the best performance across all models. However, increasing the number of linear layers to three greatly harms the model's performance. We surmise that this could be attributed to the small number of training tasks, which causes overfitting of the model at the meta-level when its meta-learning capacity is increased\cite{rajendran2020meta}. However, this does not pose any significant issues since using fewer linear layers is desirable because it decreases the computational complexity of the final model.

\begin{figure}[h]
\includegraphics{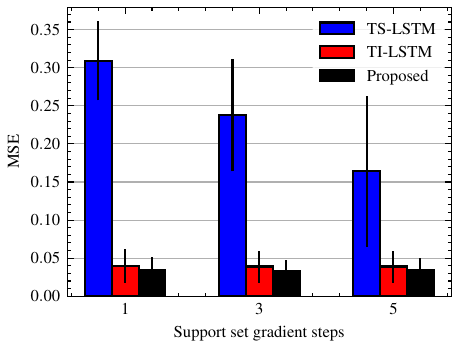}
\caption{MSE values for different numbers of support set gradient steps.}
\label{num_steps_img}
\end{figure}

\textbf{Inner Loop Steps.} For this experiment, the number of gradient steps performed within each task gradually increases to five. In the proposed model, these gradient steps refer to the number of training epochs in the inner loop optimization for each task separately. The number of gradient steps for the TI-LSTM base model refers to the number of fine-tuning epochs using the training data of a new time series. In contrast, the same variable for TS-LSTM refers to the number of training epochs using the training set of its respective time series. In Fig. \ref{num_steps_img}, the MSE values of the examined methods for varying numbers of inner loop steps are reported. TI-LSTM achieves $0.040$ for one inner loop step and $0.039$ for three and five steps, while our proposed method achieves an MSE value of $0.035$ for one and five steps and $0.033$ for three steps, improving performance by at least $10\%$ in all three cases. At the same time, its performance is only slightly affected by the number of gradient steps, demonstrating its robustness under different settings regarding this specific hyperparameter.

\begin{figure}[h]
\includegraphics{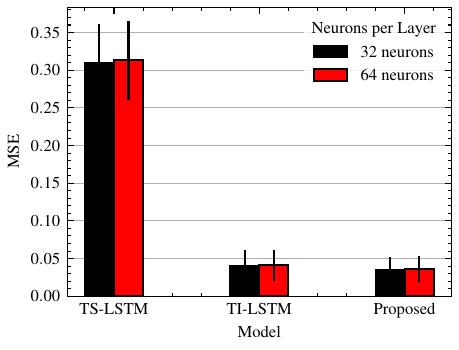}
\caption{MSE values for different numbers of neurons per layer.}
\label{num_neurons_img}
\end{figure}

\textbf{Base Learner Size.} In addition to the depth of the base learner, it would also be interesting to explore how model performance would be affected by increasing the model size. Increasing model size could augment the model's capacity, enabling better generalization to unseen data. However, a significant increase in the model's capacity could lead to overfitting. This experiment increases the number of neurons per layer from 32 to 64. Fig. \ref{num_neurons_img} depicts the change of each model's MSE when varying the number of neurons per layer from 32 to 64. Similarly to the case of varying gradient steps, our method is robust under model size variations, as evidenced by the unchanged MSE value when doubling the number of neurons per layer. At the same time, it outperforms TI-LSTM for both numbers of neurons per layer.

\section{Conclusion} \label{conclusion}
During the past few decades, load forecasting approaches have typically assumed the existence of large volumes of data for training and evaluation. However, adapting to load consumption time series with limited data is critical, allowing for seamless forecasting to newly integrated consumers and active assets while increasing robustness under equipment failures that may lead to corrupted data.

In this paper, the adaptation of a meta-learning algorithm has been proposed to solve the problem of few-shot load forecasting. By optimizing across a set of time series using gradient descent, the meta-learner can learn an optimal set of initial parameters for an LSTM base-learner. Using these initial parameters, the LSTM can adapt to each previously unseen time series by performing only a few gradient steps in its training set. This way, the LSTM model can be specifically optimized for each time series without requiring large training data. Additionally, by adopting various features of the original model, such as learnable inner loop learning rates, cosine annealing in the outer optimization loop, higher-order derivatives computation annealing, and adapting a multi-step loss, we ensure the model is stable, robust, and efficient.

In our experiments, we compare our proposed model with transfer learning and task-specific machine learning approaches and demonstrate its superiority in forecasting next week's load measurements for time series ranging from one to three months by improving performance by at least $12.5\%$. Interestingly, the model also retains state-of-the-art performance even under variations in the base learner's depth and size, the number of inner loop optimization steps, and the size of the support set. Finally, a novel metric, mean average log percentage error, is proposed, allowing for a more robust assessment of model performance by alleviating the bias introduced by the commonly used MAPE metric.

Finally, as future work, it would be worth investigating the proposed method's performance for individual consumer load forecasting, which is more challenging due to the high variances and random patterns that distinguish these time series.

\bibliographystyle{IEEEtran}
\bibliography{IEEEabrv,bibliography}

\end{document}